\documentclass{article}

     \PassOptionsToPackage{numbers, compress}{natbib}
   \usepackage[nonatbib]{neurips_2024}

\usepackage[utf8]{inputenc} % allow utf-8 input
\usepackage[T1]{fontenc}    % use 8-bit T1 fonts
\usepackage{hyperref}       % hyperlinks
\usepackage{url}            % simple URL typesetting
\usepackage{booktabs}       % professional-quality tables
\usepackage{amsfonts}       % blackboard math symbols
\usepackage{nicefrac}       % compact symbols for 1/2, etc.
\usepackage{microtype}      % microtypography
\usepackage{xcolor}         % colors

\usepackage{float}
\usepackage{graphicx}
\usepackage{url}
\usepackage{natbib}
\usepackage{amsmath}
\usepackage{algorithm}
\usepackage{algpseudocode}
\usepackage{amsfonts}
\usepackage{amssymb}
\usepackage{multirow}
\usepackage{booktabs}
\usepackage{xcolor}
\usepackage{array}
\usepackage{tabularx}
\usepackage{natbib}
\setcitestyle{square, comma, numbers,sort&compress, super}

\graphicspath{{images}}

\title{SinkLoRA: Enhanced Efficiency and  Chat Capabilities for Long-Context Large Language Models}

\author{%
  David S.~Hippocampus\thanks{Use footnote for providing further information
    about author (webpage, alternative address)---\emph{not} for acknowledging
    funding agencies.} \\
  Department of Computer Science\\
  Cranberry-Lemon University\\
  Pittsburgh, PA 15213 \\
  \texttt{hippo@cs.cranberry-lemon.edu} \\
}

\begin{document}

\maketitle

\begin{abstract}
Extending the functionality of the Transformer model to accommodate longer sequence lengths has become a critical challenge. This extension is crucial not only for improving tasks such as language translation and long-context processing but also for enabling novel applications like chatbots, code generation, and multimedia content creation. The primary obstacle is the self-attention mechanism, which scales quadratically with sequence length in terms of computation time and memory requirements.
LongLoRA proposed shifted sparse attention (S\(^2\)-Attn), effectively enabling context extension and leading to non-trivial computation savings with similar performance to fine-tuning with vanilla attention. However, LongLoRA is still not as efficient as vanilla attention, reaching only 39\% of the perplexity improvement compared to full attention. This inefficiency is due to the cyclic shift applied within different attention head patterns, causing either chaos in the attention head structure or unnecessary information exchange between token groups.
To address these issues, We propose \textbf{SinkLoRA}, which features better work partitioning. Specifically, (1) we developed SF-Attn with a segmentation and reassembly algorithm to proportionally return cyclically shifted groups of attention heads to their un-shifted state together with global attention of "sink attention tokens", achieving 92\% of the perplexity improvement compared to full attention after fine tuning, and (2) applied a SOTA KV cache compression algorithm H$_2$O to accelerate inference. Furthermore, We conducted supervised fine-tuning with SinkLoRA using a self collected LongAlpaca-plus dataset. All our code, models, datasets, and demos are available at \url{https://github.com/Dexter-GT-86/SinkLoRA}.
\end{abstract}

\begin{figure}
\centering
\includegraphics[width=\linewidth]{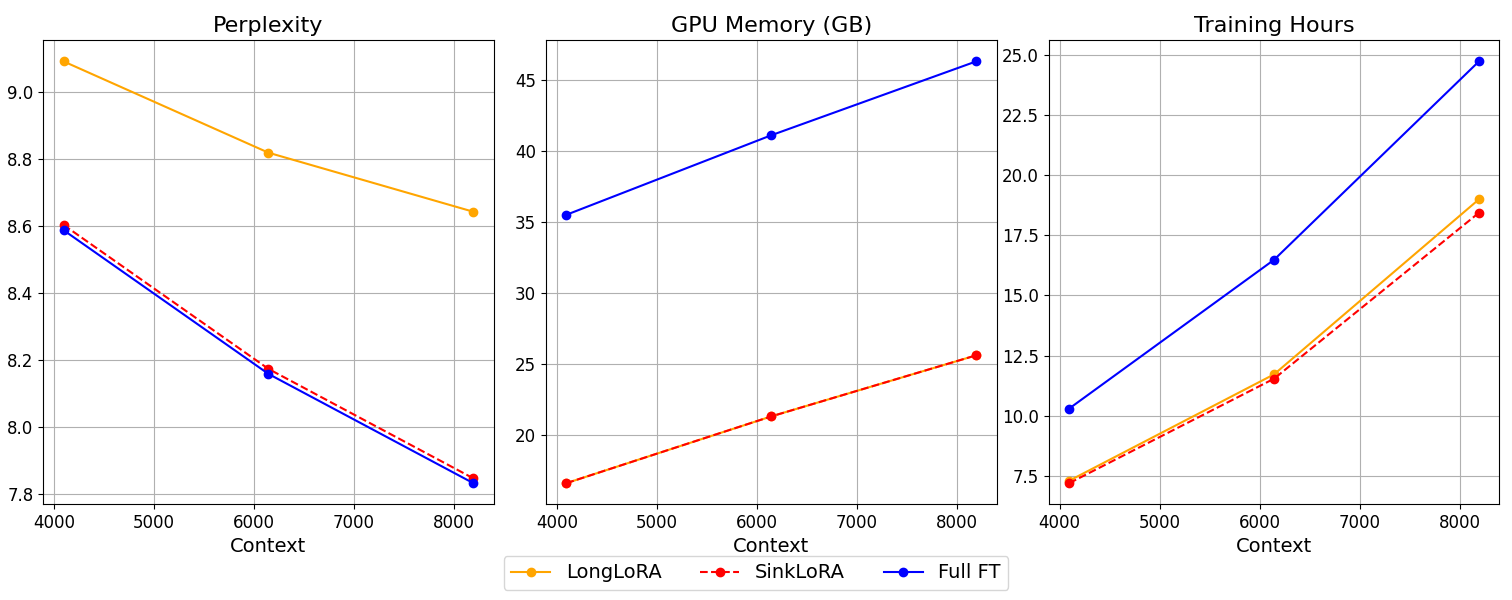}
\caption{\label{fig:figure1} Evaluation of SinkLoRA in bridging the accuracy gap between sparse shifted attention and full attention during supervised fine-tuning, while maintaining the memory efficiency of LongLoRA, which utilizes 1.8 times less memory compared to full fine-tuning. Furthermore, SinkLoRA retains the training speed of LongLoRA, being 1.8 times faster than full fine-tuning, due to the implementation of Sink Fixed Attention. The Llama2-7B models\cite{touvron2023llama} are fine-tuned to various context lengths using Flash-Attention2 \cite{dao2023flashattention} and DeepSpeed stage 2 \cite{rasley2020deepspeed}, and are evaluated on the proof-pile test set \cite{azerbayevproof} in terms of perplexity.}
\end{figure}

\begin{figure}
\centering
\includegraphics[width=1\linewidth]{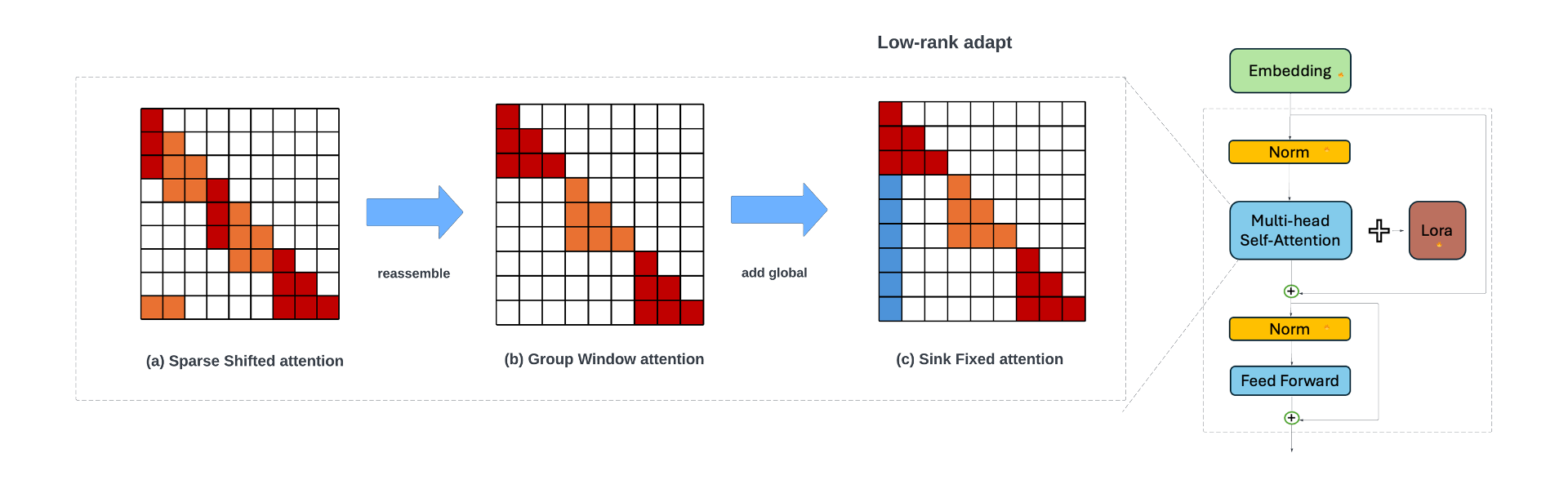}
\caption{\label{fig:figure2}Overview of the SinkLoRA fine-tuning process, incorporating Sink Fixed Attention (SF-Attn). Panels (a), (b), and (c) depict the procedure to convert Sparse Shifted Attention into Short Window Attention and subsequently into Sink Fixed Attention. This conversion is executed in two stages: reassembly and making the initial tokens global. In addition to optimizing the LoRA weights within linear layers, SinkLoRA also enables training of the embedding and normalization layers, consistent with the methodology employed in LongLoRA.}
\end{figure}
% (This figure is adapted from \cite{chen2023longlora}.) 

\begin{figure}
\centering
\includegraphics[width=1\linewidth]{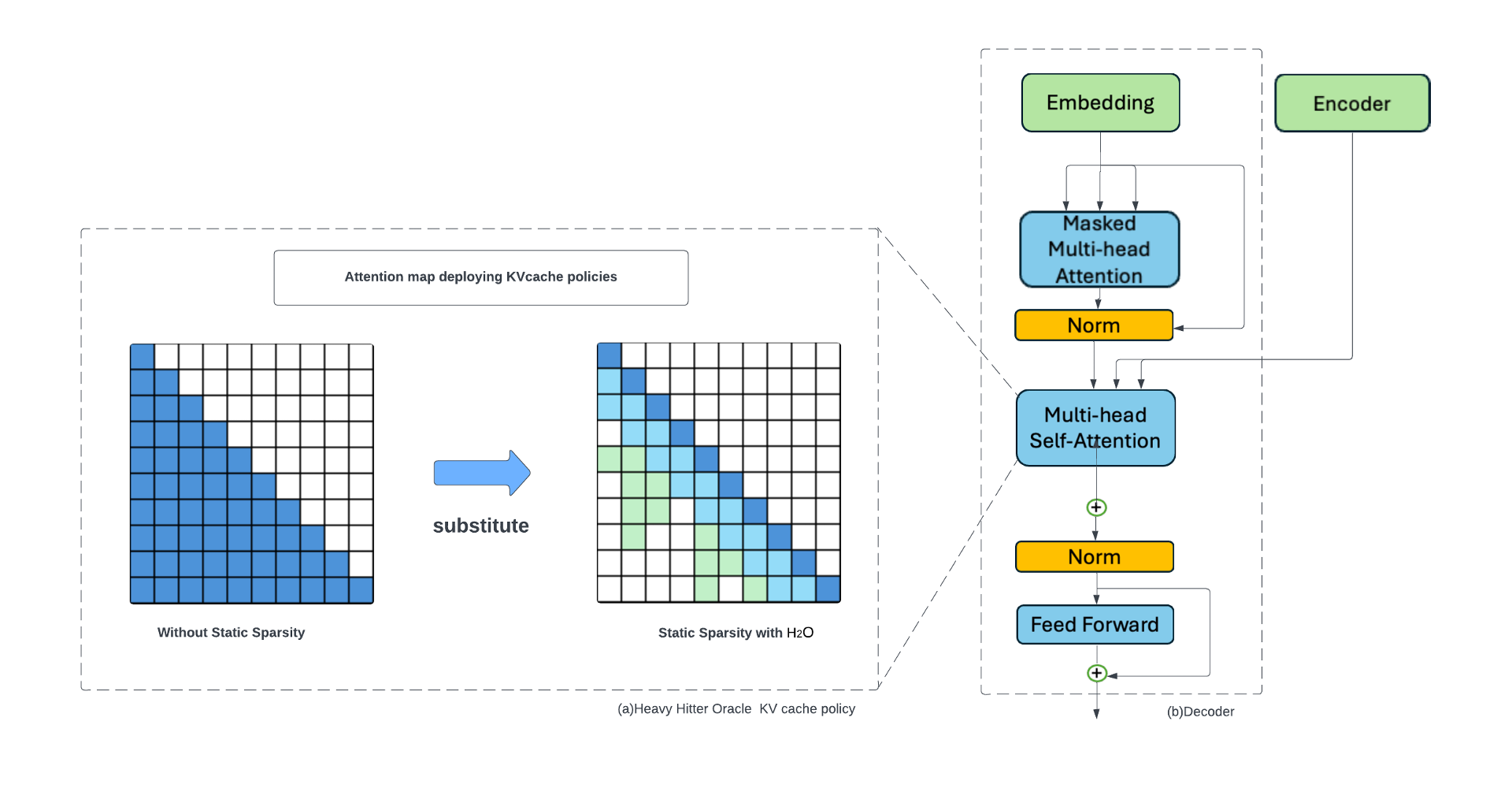}
\caption{\label{fig:figure3}Overview of the SinkLoRA inference process. Unlike LongLoRA, which retains the original standard self-attention during inference, SinkLoRA implements an optional KV cache compression method, H\(^2\)O \cite{zhang2024h2o}. This extension enhances inference speed without significantly compromising performance.}
\end{figure}

% (This figure is adapted from \cite{zhang2024h2o}.) 

\section{Introduction}

Enhancing the functionality of Transformer models to handle longer sequence lengths has become crucial for numerous applications, including language translation, long-context processing, chatbots, code generation, and multimedia content creation. The primary challenge lies in the self-attention mechanism, which scales quadratically with sequence length, leading to substantial computational time and memory requirements \cite{beltagy2020longformer, zaheer2020big, kitaev2020reformer}. To address this challenge, several approaches have been proposed. Longformer and BigBird utilize combinations of local, global, and sparse attention mechanisms to manage long contexts, reducing complexity to O(n) \cite{beltagy2020longformer, zaheer2020big}. Reformer introduces locality-sensitive hashing (LSH) to approximate attention by hashing similar tokens into the same buckets, thereby reducing computational complexity \cite{kitaev2020reformer}. LSG Attention combines local, sparse, and global attention to effectively handle long contexts while minimizing computational overhead \cite{condevaux2023lsg}.

Despite these advancements, managing long-context interactions in practical applications remains a significant challenge. Recent work, such as LongLoRA, extends the context window of LLaMA2 from 4096 to 32768 tokens using Position Interpolation without substantial GPU or TPU resources \cite{chen2023longlora}. However, LongLoRA’s efficiency is limited, achieving only 39\% of the perplexity improvement compared to full attention due to chaotic attention head structures and unnecessary information exchange between token groups.

To address these issues, we propose \textbf{SinkLoRA}, which offers better work partitioning. This includes the development of Sink Fixed Attention (SF-Attn), a segmentation and reassembly algorithm that, along with the global attention of ``sink attention tokens,'' achieves 92\% of the perplexity improvement of full attention after fine-tuning. Additionally, we apply a state-of-the-art KV cache compression algorithm, Heavy Hitter Oracle (H$_2$O), to accelerate inference \cite{zhang2024h2o, ge2023model, liu2024scissorhands}.

We further enhanced SinkLoRA through supervised fine-tuning using our self-collected LongAlpaca-Plus dataset, comprising 28,000 entries from various sources, including Natural Questions, RedPajama \cite{togethercomputer2023redpajama}, Book Summarization, and LongQA \cite{chen2023longlora}, ensuring a diverse and comprehensive collection for long instruction tuning.

In summary, the contributions of our work are as follows:
\begin{itemize}
    \item We present SinkLoRA, a memory-efficient and effective method to extend the context length of LLaMA2 and LLaMA3, representing a complete update of LongLoRA. This method improves fine-tuning efficiency and offers a flexible deployment inference strategy.
    \item We introduce SF-Attn, a fine-tuning method that combines a segmentation \& reassembly algorithm and global attention. This method is easy to implement, accurate, and memory-efficient, without increasing computational complexity. By directly modifying the attention pattern, SF-Attn effectively redistributes attention scores, reducing the undue emphasis on initial tokens across different token groups.
    \item We achieve efficient deployment of computationally intensive large language models (LLMs) in production environments by using the Heavy Hitter Oracle (H$_2$O) KV caching method. This method stores the key-value states of previously generated tokens, significantly reducing the need for repetitive computations and thus lowering latency in autoregressive generation. This enhancement allows for a more flexible and efficient inference strategy, reducing computational overhead while maintaining model performance.
    \item Our SinkLoRA performs favorably against state-of-the-art methods. We evaluate its performance on the PG19, Proof-pile, and LongBench datasets, demonstrating its effectiveness. Specifically, for LLaMA2 7B, SinkLoRA outperforms LongLoRA and is competitive with LongChat-13B.
   
\end{itemize}

\subsection{Motivation for the Research}

\textbf{Motivation 1: Elevating Attention Scores for Initial Tokens} \\
Prior studies have demonstrated the Attention Sink phenomenon, where certain tokens, typically the initial tokens in a sequence, receive disproportionately high attention scores during the model’s computation \cite{xiao2023efficient}. This often occurs because these tokens are visible to all subsequent tokens, leading to significant attention even when they lack semantic importance, particularly in autoregressive language models \cite{sandal2024zero}.

The Sparse Shifted Attention mechanism implemented in LongLoRA \cite{chen2023longlora} attempts to address this by shifting the high attention scores from these initial tokens to other tokens that previously received lower attention. This shift reduces the overemphasis on initial tokens. To further improve this, we need to develop a method that directly modifies the attention pattern. By applying this technique, we can effectively redistribute attention scores, thereby reducing the undue emphasis on initial tokens across different token groups.

\textbf{Motivation 2: Maintaining Initial Tokens During Fine-Tuning} \\
The concept of attention sinks is also utilized in Streaming LLM \cite{xiao2023efficient} to improve the model’s handling of long texts. By retaining the Key-Value (KV) pairs of a few initial tokens (attention sinks) along with the most recent tokens, the model ensures stable attention scores and performance even for extended sequences. Inspired by this approach, we aim to carry this mindset from training into inference. Our research aims to modify the fine-tuning process so that initial tokens attend to all other tokens, thereby accumulating more attention scores and enhancing the model’s capacity to handle long sequences.

\textbf{Motivation 3: Flexible Deployment of Inference Strategy} \\
Efficient deployment of computationally intensive large language models (LLMs) in production environments often relies on Key-Value (KV) caching \cite{ge2023model}. KV caching stores the key-value states of previously generated tokens, significantly reducing the need for repetitive computations and thus lowering latency in autoregressive generation. However, LongLoRA \cite{chen2023longlora} retains only the original standard self-attention mechanism during inference. To address this limitation, it is necessary to apply an optional KV cache function. This enhancement allows for a more flexible and efficient inference strategy, reducing computational overhead while maintaining model performance.

\begin{figure}
\centering
\includegraphics[width=1\linewidth]{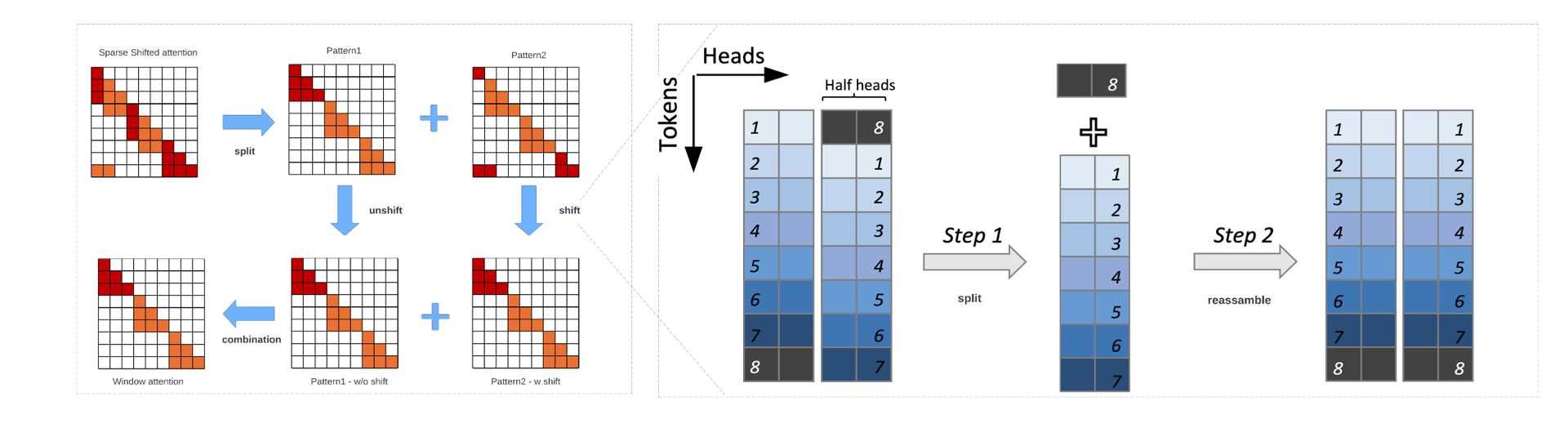}
\caption{\label{fig:figure4} Illustration of the Segmentation and Reassembly process in SF-Attn. The process involves three steps: (1) Splitting features along the head dimension into two chunks: one shifted and one unshifted. (2) Splitting tokens, where the tokens belonging to the shifted chunk are shifted by half of the group size, and reassembling them at the tail of the tokens to match the unshifted chunk. (3) Combining the two chunks of tokens together. This figure is adapted from \cite{chen2023longlora}.}
\end{figure}

\section{Related Work}

\subsection{Long-context Transformers}

The primary obstacle in scaling Transformer models to handle longer sequence lengths lies in the self-attention mechanism, which scales quadratically with sequence length in terms of computation time and memory requirements. This quadratic computational burden has prompted significant research efforts focused on developing more efficient sparse Transformer models. Notable examples include Longformer \cite{beltagy2020longformer} and BigBird \cite{zaheer2020big}, which utilize a combination of local, global, and sparse attention mechanisms to manage long contexts, thereby reducing the complexity to O(n). These models achieve a balance between maintaining sufficient context for understanding while managing computational load. For achieving complexity of O(n log n), several approaches have been proposed. Fixed Window Attention \cite{child2019generating} employs a fixed-size window for attention, which confines the attention computation to a limited context window. Reformer \cite{kitaev2020reformer} introduces locality-sensitive hashing (LSH) to approximate attention by hashing similar tokens into the same buckets, thus reducing the computational complexity. LSG Attention \cite{condevaux2023lsg}, adapted from BigBird, combines local, sparse, and global attention to effectively handle long contexts while minimizing computational overhead. Equipping Transformer \cite{yang2024equipping} proposes a novel reading strategy termed random access, which enables Transformers to efficiently process long documents without needing to examine every token. This method shows promising results across pretraining, fine-tuning, and inference phases, demonstrating its efficacy in handling extended contexts. Despite these advancements, the ability of these methods to manage long-context conversations, such as those required in chat applications, remains limited. This highlights an ongoing challenge in enhancing the context-handling capabilities of Transformer models for interactive and real-time applications. 

\subsection{Long-context LLMs}

Recent advancements in Large Language Models (LLMs) have significantly extended their capabilities, including handling long-context inputs. Math Word Problems (MWPs) have demonstrated notable performance in solving mathematical questions using LLMs \cite{satpute2024can}. Moreover, leveraging LLMs for SQL querying has shown promise in optimizing resource allocation, though it remains less efficient than traditional relational databases \cite{zhang2024can}. LongLoRA \cite{chen2023longlora}, employing Position Interpolation \cite{chen2023extending}, has successfully extended the context window of Llama 2 from 4096 to 32768 tokens without requiring substantial GPU or TPU resources. Meta's Llama 3, featuring up to 70 billion parameters, represents a significant advancement in open-source LLMs, offering enhancements in computational efficiency, trust and safety tools, and collaborations with major platforms \cite{meta2024llama3}. Open-source models such as BLOOM \cite{le2023bloom}, OPT \cite{iyer2022xi}, and Falcon \cite{penedo2023refinedweb} continue to challenge proprietary models, although models like Vicuna \cite{peng2023instruction} and Alpaca \cite{andermatt2023uzh_pandas} still lag behind their closed-source counterparts in certain aspects. Despite these advancements, effectively managing long-context interactions remains a significant challenge, necessitating ongoing research and development to address the complexities in long-context LLM applications.

\subsection{KV-Cache Compression}
Compressing the size of KV cache is more difficult than reducing the size of weights because they are more sensitive and dependent on model inputs. A cost-effective method for KV cache compression is token dropping \cite{liu2024scissorhands, zhang2024h2o, ge2023model}, which establishes an importance policy to retain significant KVs and remove insignificant ones. Jiang et al. \cite{jiang2023mistral} and Xiao et al. \cite{xiao2023efficient} suggest preserving tokens that are local to the current sequence position, as these are crucial for generation. For example, H$_2$O \cite{zhang2024h2o} and FastGen \cite{ge2023model} proposed reducing KV cache size by dropping tokens based on their attention scores.  Similarly, SparQ \cite{ribar2023sparq} drops tokens according to attention score sparsity and also considers the error in the pruned value cache.

These pruning methods are generally effective for summarization tasks and zero-shot inference. Zhang et al. \cite{zhang2024h2o} and Liu et al. \cite{liu2024scissorhands} recommend identifying a small set of influential tokens, termed heavy-hitters, to better maintain generation quality. Ge et al. \cite{ge2023model} have empirically demonstrated that different attention heads prioritize different tokens and have developed an adaptive importance policy for evicting KVs. However, these approaches can cause significant issues since the context contained in the evicted KVs is entirely discarded.

\begin{table}[H]
	\centering
	\fontsize{8}{11}\selectfont    %{字体尺寸}{行距}
	\caption{Effectiveness of SF-Attn under different context lengths. ‘Short’ refers to 1/4 of the target context length, while ‘Long’ equals the target context length. 'Shift' indicates token exchange between different groups of tokens, whereas 'Global' means initial tokens attend to all tokens. Models are supervised fine-tuned on a Llama2 \cite{touvron2023llama} model with 7B parameters on the LongAlpaca-plus dataset. Results are evaluated in terms of perplexity on the PG19 \cite{rae2019compressive} validation split.}
 \vspace{0.5cm}

	\begin{tabular}{ccccc}
		\toprule
		Setting & Training & \multicolumn{3}{c}{Target Context Length} \\
		\cmidrule(lr){3-5}
		& & 4096 & 6144 & 8192 \\
		\midrule
		Full Attention & Long & 8.59 & 8.16 & 7.8 \\
		Sparse Shifted Attention & Short \& Shift & 9.09 & 8.82 & 8.6 \\
		Sink Fixed Attention & Short \& Global & \textbf{8.60} & \textbf{8.17} & \textbf{7.85} \\
		\bottomrule
	\end{tabular}
	\label{tab:table1}
\end{table}

\section{Our Method: SinkLoRA}

\subsection{Background}

\textbf{LongLoRA}.
LongLoRA, introduced by Chen et al. (2023) \cite{chen2023longlora}, is an innovative fine-tuning methodology aimed at extending the context window sizes of large language models (LLMs) efficiently. Leveraging Position Interpolation \cite{chen2023extending}, LongLoRA builds upon the low-rank adaptations introduced by LoRA \cite{hu2021lora} and incorporates Shifted Sparse Attention (S\(^2\)-Attn) \cite{chen2023longlora}. Inspired by the Swin Transformer \cite{liu2021swin}, S\(^2\)-Attn manages extended contexts by partitioning the total context into multiple groups and performing attention computations independently within each group. To ensure continuity, it shifts half of the attention heads by half the group size. This method effectively simulates long-context operations using short attention spans during training, allowing for the handling of significantly larger contexts without a substantial increase in computational overhead. Furthermore, LongLoRA is designed to be compatible with existing LLM optimization techniques and infrastructures, such as Flash-Attention2 \cite{dao2022flashattention, dao2023flashattention}, thereby enhancing its usability without requiring major system modifications. The approach also emphasizes efficient parameter utilization, necessitating minimal adjustments to the learnable embedding and normalization layers, which represent a small portion of the overall model parameters. This efficiency is crucial for scaling up to larger models while maintaining the practicality of LongLoRA for enhancing performance in tasks that demand deep contextual understanding.

\textbf{Attention Sink}.
In autoregressive large language models (LLMs), an intriguing phenomenon known as the “attention sink” \cite{xiao2023efficient} is observed, where initial tokens receive a disproportionate amount of attention scores, regardless of their semantic relevance to the task. These initial tokens, although lacking significant semantic importance, tend to accumulate high attention scores. This phenomenon arises due to the nature of the Softmax function used in the attention mechanism, which ensures that attention scores across all tokens sum to one. In situations where few tokens are strongly related to the current query, the model redistributes attention to available tokens, often defaulting to the initial ones. Given the autoregressive nature of LLMs, where each token predicts the next in sequence, initial tokens are consistently visible to nearly all subsequent tokens, inherently training them to become preferred targets for attention, thus acting as “attention sinks.” This insight highlights a fundamental challenge in the attention mechanism of autoregressive models and suggests areas for improving attention distribution strategies in language modeling.

 \textbf{Heavy-Hitter Oracle}.
The Heavy-Hitter Oracle (H$_2$O) \cite{zhang2024h2o} is a SOTA method
that addresses the challenge of reducing the memory footprint of the KV cache in language models. This method is based on the observation that a small subset of tokens, termed “Heavy Hitters” (H2), significantly contribute to the overall attention scores in language models. Analysis shows that these H2 tokens frequently co-occur within the text, making their presence a natural outcome of the text’s structure. Experiments indicate that removing these tokens can significantly impair model performance, underscoring their importance. The H$_2$O approach combines this understanding with a dynamic cache eviction policy that optimally balances the retention of recent tokens and crucial H2 tokens. This strategy ensures efficient memory use while maintaining the computational dynamics essential for robust model performance.

\newcommand{\coloredcomment}[1]{\textcolor{green}{#1}}

\begin{algorithm}
\caption{Segmentation and Reassembly Algorithm of q, k in PyTorch-like style}
\begin{algorithmic}[1]
\Function{SegmentationAndReassembly}{$qkv, shift\_size$}
    \State \textbf{Input:} $qkv$: input tensor of shape $(B, N, 3, H, D)$; $shift\_size$: size of the shift
    \State \textbf{Output:} $output$: tensor after segmentation and reassembly
    \State $B, N, _, H, D \gets \text{shape of } qkv$
    \State $G \gets \text{int}(N / shift\_size)$
    \State $shift \gets \text{int}(G / (2 / shift\_size))$
    
    \State \textcolor{teal}{\# Perform the operation on the second half of the heads}
    \State $last\_half \gets qkv[:, :, :, H // 2:]$
    
    \State \textcolor{teal}{\# Split the tensor into two parts: one to stay and one to be moved}
    \State $to\_wrap \gets last\_half[:, :, :, :shift]$
    \State $to\_move \gets last\_half[:, :, :, shift:]$
    
    \State \textcolor{teal}{\# Concatenate the wrap-around part with the moved part}
    \State $new\_last\_half \gets \text{torch.cat}((to\_move, to\_wrap), \text{dim}=3)$
    
    \State \textcolor{teal}{\# Replace the original last half with the new configuration}
    \State $qkv[:, :, :, H // 2:] \gets new\_last\_half$
    
    \State \textcolor{teal}{\# Standard self-attention function}
    \State $out \gets \text{self\_attn}(qkv)$
    
    \State \textcolor{teal}{\# Split the output tensor into two parts: one to stay and one to be moved back}
    \State $to\_wrap\_back \gets out[:, :, :, :shift]$
    \State $to\_move\_back \gets out[:, :, :, shift:]$
    
    \State \textcolor{teal}{\# Concatenate the wrap-around part with the moved part}
    \State $new\_out \gets \text{torch.cat}((to\_move\_back, to\_wrap\_back), \text{dim}=3)$
    
    \State \textcolor{teal}{\# Replace the original output with the new configuration}
    \State $out \gets new\_out$
    
    \State \textcolor{teal}{\# Feedforward operation}
    \State $output \gets \text{feedforward}(out)$
    
    \State \Return $output$
\EndFunction
\end{algorithmic}
\label{algorithm1}
\end{algorithm}

\subsection{APPLYING Sink Fixed Attention to Improve Fine Tune Progress}

\subsubsection{Pilot study}
To confirm our approach, we conducted a pilot study to evaluate the effectiveness of different attention mechanisms under various context lengths.

In Table \textcolor{red}{\ref{tab:table1}}, we establish a standard baseline by training and testing with full attention and fine-tuning, which consistently delivers good quality across various context lengths. Our first trial involved training with short attention(Sparse Shifted Attention), represented by the pattern in Figure 2(a).

In this trial, given the high computational cost associated with self-attention modules for long contexts, to address this issue, we introduced Sink Fixed Attention (SF-Attn) reassembly into group window attention, as shown in Figure \textcolor{red}{\ref{fig:figure2}} (b). This stage ensures better continuity and interaction between tokens by reassembling the attention pattern, depicted in Figure \textcolor{red}{\ref{fig:figure2}} (c). This method enhances the attention mechanism by making initial tokens attend to all tokens, thus providing global attention. As a result, SF-Attn demonstrates significant improvements over Sparse Shifted Attention, reducing perplexity notably across all target context lengths, especially at 8192 tokens.

The findings from this pilot study indicate that SF-Attn, through reassembly and global attention adjustments, effectively balances performance and computational efficiency. This makes it a promising solution for extending the context length of large language models in practical applications. there is no information exchange between different groups.

The detail of SF-Attn will be show in Section 3.2.2

\subsubsection{Sink Fixed Attention}

Sink Fixed Attention includes two main parts: Segmentation \& Reassembly Algorithm, Global Attention.

\textbf{Segmentation \& Reassembly Algorithm}

As shown in \textcolor{red}{\ref{fig:figure4}}, the input tensor is divided into two halves. The second half is split into two parts: one part stays in place while the other part is shifted and wrapped around. These parts are then combined back together in a new order. This reassembled tensor is processed through the standard self-attention mechanism. Finally, the output tensor undergoes a similar splitting, shifting, and recombining process before passing through a feed forward operation to produce the final output. This approach simplifies the attention calculation and improves the model’s performance by focusing on key areas of the input. This makes the sparse shifted attention similar to the fixed window attention. We provide a PyTorch-style code in Algorithm \textcolor{red}{\ref{algorithm1}}.

\textbf{Global Attention}

We choose the first four initial tokens as "sink attention tokens". When training Llama-2-7B and Llama-3-8B models, this setting follows the structure of StreamingLLM \cite{xiao2023efficient}. Using the global attention strategy from Longformer and BigBird \cite{beltagy2020longformer,zaheer2020big}, we make the sink attention tokens attend to all tokens across the sequence, and all tokens in the sequence attend to the sink attention tokens. Specifically, after the Segmentation \& Reassembly Algorithm, we obtain the attention scores map of all the tokens in the currently fixed window as a shifted back version.

Add \( g \) new tokens (sink attention tokens) to the current sequence. Correspondingly, construct a new adjacency matrix \( B \in [0, 1]^{(N+g) \times (N+g)} \). For \( i \in \{1, 2, \ldots, g\} \) and \( B(g+i, g+j) = A(i, j) \forall i, j \in \{1, \ldots, N\} \), it holds that \( B(i, :) = 1 \) and \( B(:, i) = 1 \).

By doing this, we keep the sink attention tokens attending in attention score calculation before going inside hidden layers.
Since the number of such tokens is small relative to and independent of \( n \), the complexity of the combined local and global attention is still \( O(n \log n) \).
% as global token

% Accordingly, we add “global attention” on few pre-selected input locations. Importantly, we make this attention operation symmetric: that is, a token with a global attention attends to all tokens across the sequence, and all tokens in the sequence attend to it. Fig. 2d shows an example of a sliding window attention with global attention at a few tokens at custom locations. For example for classification, global attention is used for the [CLS] token while in QA global attention is pro- vided on all question tokens. Since the number of such tokens is small relative to and independent of n the complexity of the combined local and global attention is still O(n). While specifying global attention is task specific, it is a easy way to add in- ductive bias to the model’s attention, and it is much simpler than existing task specific approaches that use complex architecture to combine information across smaller input chunks.

\subsection{Applying KV Cache Algorithm to Accelerate Inference Process}
In this section, we apply the H$_2$O algorithm to LongLoRA. The LongLoRA codebase\cite{chen2023longlora} only provides a full attention solution for the inference stage. We incorporate the H$_2$O eviction system \cite{zhang2024h2o} by creating a new class to manage the KV cache values in memory, thus optimizing the inference process. This integration is illustrated in Figure \textcolor{red}{\ref{fig:figure3}}.

% \subsection{Collecting a more diverse supervised fine-tune dataset }
\begin{table}[H]
	\centering
	\fontsize{8}{11}\selectfont    %{字体尺寸}{行距}
	\caption{Perplexity evaluation on the proof-pile \cite{rae2019compressive} test split. S\(^2\)-Attn: Shifted Sparse Attention. SF-Attn: Sink Fixed Attention. We supervised fine-tune Llama2 \cite{touvron2023llama} and Llama3 \cite{meta2024llama3} in 7B and 8B model sizes on the LongAlpaca dataset \cite{chen2023longlora} under 8k-32k context lengths. Our method achieves comparable performance to the full attention or full FT baselines, with better efficiency. We use the same training settings as the model evaluated on PG19 \cite{rae2019compressive} as introduced in Section 4.1 in the evaluation. }
 \vspace{0.5cm}

	\begin{tabular}{cccccc}
		\toprule
		\multirow{2}{*}{Setting} & \multirow{2}{*}{Training} & \multirow{2}{*}{Attention} & \multicolumn{3}{c}{Target Context Length} \\
		\cmidrule(lr){4-6}
		& & & 4096 & 6144 & 8192 \\
		\midrule
		\multirow{3}{*}{7B-Llama2}
		& & Full-Attn & 8.59 & 8.16 & 7.83 \\
		& 8192 & S\(^2\)-Attn & 9.10 & 8.82 & 8.6 \\
		& & SF-Attn & \textbf{8.60} & \textbf{8.17} & \textbf{7.85} \\
		\midrule
		\multirow{3}{*}{8B-Llama3}
		& & Full-Attn & 5.10 & 4.85 & 4.65 \\
		& 8192 & S\(^2\)-Attn & 5.40 & 5.24 & 5.13 \\
		& & SF-Attn & \textbf{5.11} & \textbf{4.86} & \textbf{4.66} \\
		\bottomrule
	\end{tabular}\vspace{0cm}
	\label{tab:table2}
\end{table}

% \usepackage{multirow}
\begin{table}[H]
\centering
\caption{Maximum context length that we can fine-tune for various model sizes on a single 2× A100 machine. We use the same training and evaluation settings as in Table \textcolor{red}{\ref{tab:table2}}. Flash-Attention2 \cite{dao2023flashattention} and DeepSpeed \cite{rasley2020deepspeed} are used in stage 3 during fine-tuning. With LongLoRA, the maximum context length for 7B of Llama-2 and 8B of llama-3 models are 16k. Evaluation on PG19 \cite{rae2019compressive} is discussed in Section 4.1 in the evaluation.}
\vspace{0.5cm}

\begin{tabular}{|l|l|lllll|}
\hline
\multirow{2}{*}{Model} & \multirow{2}{*}{\begin{tabular}[c]{@{}l@{}}Training\\ Context Length\end{tabular}} & \multicolumn{5}{l|}{Evaluation Context Length} \\ \cline{3-7}
                       &                                                                                     & \multicolumn{1}{l|}{2048} & \multicolumn{1}{l|}{4096} & \multicolumn{1}{l|}{8192} & \multicolumn{1}{l|}{12288} & 16384 \\ \hline
7B-Llama-2             & 16384                                                                               & \multicolumn{1}{l|}{8.73} & \multicolumn{1}{l|}{8.55} & \multicolumn{1}{l|}{8.30} & \multicolumn{1}{l|}{8.13}  &\textbf{8.05}  \\ \hline
8B-Llama-3             & 16384                                                                               & \multicolumn{1}{l|}{5.11} & \multicolumn{1}{l|}{5.02} & \multicolumn{1}{l|}{4.89} & \multicolumn{1}{l|}{4.77}  & \textbf{4.72}  \\ \hline
\end{tabular}\vspace{0cm}
\label{tab:table3}
\end{table}
\vspace{0.5cm}

\begin{table}[H]
    \centering
    \caption{Topic retrieval evaluation with LongChat \cite{li2023long}. We compare our model to other open-source long-context LLMs. This task involves retrieving target topics from very long conversations with around 3k, 6k, 10k, 13k, and 16k context lengths. As some questions in the evaluation set are longer than 16k, our model is fine-tuned on Llama3 8B. It achieves comparable performance to the state-of-the-art LongChat-13B \cite{li2023long} with a lower fine-tuning cost.}
    \vspace{0.5cm}

    \begin{tabular}{lccccc}
        \toprule
        \textbf{Evaluation Context} & \textbf{3k} & \textbf{6k} & \textbf{10k} & \textbf{13k} & \textbf{16k} \\
        \midrule
        ChatGLM2-6B \cite{du2021glm} & 0.88 & 0.46 & 0.02 & 0.02 & 0.02 \\
        MPT-30B-chat \cite{team2023a} & 0.96 & 1.00 & 0.76 & - & - \\
        MPT-7B-storywriter \cite{team2023b} & 0.46 & 0.46 & 0.28 & 0.34 & 0.36 \\
        LongChat-13B \cite{li2023long} & \textbf{1.00} & \textbf{1.00} & \textbf{1.00} & 0.98 & 0.90 \\
        LongLoRA-13B \cite{chen2023longlora} & \textbf{1.00} & 0.98 & 0.98 & \textbf{0.98} & \textbf{0.94} \\
        \cmidrule(lr){1-6}
        Ours-8B & \textbf{1.00} & \textbf{1.00} & \textbf{1.00} & \textbf{0.98} & \textbf{0.96} \\
        \bottomrule
    \end{tabular}
    \label{tab:table4}
\end{table}

\section{Experiment}
\subsection{Experimental Settings}

\subsubsection{ENVIRONMENTS}
All our experiments were conducted on a machine equipped with 2× A100 GPUs. We utilize PyTorch \cite{paszke2017automatic} as the primary framework for training all models, integrating DeepSpeed \cite{rasley2020deepspeed} and Flash-Attention2 \cite{dao2023flashattention} to optimize performance. By default, DeepSpeed was employed in stage 2, while stage 3 was reserved for experiments involving the maximum context length. Gradient checkpointing, a standard technique in the Peft codebase \cite{ding2023parameter}, was used to manage memory efficiently. It is worth noting that while our experiments predominantly utilized 2× A100 GPUs, the use of RTX 4090 GPUs is also feasible for certain tasks, such as fine-tuning 7B models to an 8192 context size.

\subsubsection{Models} 
We extend the pre-trained 7B chat version of Llama2\cite{touvron2023llama}  and 8B of Llama3 \cite{meta2024llama3}models. 

\subsubsection{Training Procedure}
To train the model, we use the Llama-2 and llama-3 model with bf16 precision enabled and a maximum sequence length base on requirement. Flash attention \cite{dao2023flashattention}is utilized for efficient computation. Low-rank training is enabled, and the model undergoes three epochs of training with a batch size of 1 for training and 2 for evaluation per device. Gradient accumulation is set to 1 step. The evaluation strategy is disabled, while the model is saved every 512 steps with a total limit of 2 saved checkpoints. The learning rate is set to 2e-5 with no weight decay, and a warmup phase of 20 steps is included. The learning rate follows a constant schedule with warmup. Logs are recorded every step, and the training process is optimized using DeepSpeed configuration (stage2.json)\cite{rasley2020deepspeed} with tf32 precision enabled.

\subsubsection{Datasets} 
The “LongAlpaca-Plus” dataset, containing 28,000 entries, is structured with a significant focus on different sources which is a update version of LongAlpaca \cite{chen2023longlora}. Natural Questions make up 44\% of the dataset, totaling 12,320 entries, sourced from a subset of Natural Questions data. RedPajama contributes 27\% of the dataset with 7,560 entries, also derived from its specific subset. Book Summarization comprises 18\% of the dataset with 5,040 entries, and LongQA represents the smallest portion at 11\%, with 3,080 entries. This distribution ensures a diverse and comprehensive dataset for long instruction tuning tasks. The Overview of LongAlpaca-Plus dataset is provided in Section \textcolor{red}{\ref{sec:B}}.

\subsubsection{Metrics}

\textbf{Perplexity}
Perplexity, a crucial metric in natural language processing (NLP), quantifies a language model's predictive uncertainty, being inversely related to the probability assigned to the actual sequence of words. Originating from information theory, perplexity measures the efficiency of a communication system, and in NLP, it reflects how well a language model understands language patterns \cite{jelinek1977perplexity}. Mathematically, it is defined as 

\begin{equation}
\text{Perplexity} = e^{-\frac{1}{N} \sum_{i=1}^{N} \log p(w_i \mid w_{i-1}, \ldots, w_1)}
\end{equation}
 where p(x) is the probability distribution over all possible sequences x. Employed extensively in evaluating language models, from early n-gram approaches to advanced neural architectures, perplexity aids in comparing model performances, tuning during development, and enhancing tasks like machine translation and text generation \cite{jelinek1977perplexity}, \cite{finkelstein1996chess}. As a quantitative measure, it offers insights into model improvements and the effectiveness of different configurations, making it an indispensable tool in language model evaluation \cite{garofolo1999spoken}.

\textbf{Passkey Retrieval}
We using the similar of format provide in LongLoRA\cite{chen2023longlora} for passkey retrieval. The document adheres to this structure:

Within a large amount of extraneous text, crucial information is concealed.
Identify and remember this key information, as you will be tested on it.

Here is an example document structure:

-----------------------------------------------------------------------------------------------------------------------

The flowers are blooming. The trees are tall. The river flows. Just keep going. Onward and upward. (repeated M times)

Critical Note: The passkey is 84729. Remember this number. 84729 is the passkey.

The flowers are blooming. The trees are tall. The river flows. Just keep going. Onward and upward. (repeated N times)

What is the passkey? The passkey is…
-----------------------------------------------------------------------------------------------------------------------
The length of the document varies based on the values of M and N. The passkey number, such as 84729, is randomly generated and changes with each test.

 \textbf{LongBench} 
 LongBench \cite{bai2023longbench} is a bilingual, multitask benchmark designed to evaluate the long context understanding capabilities of large language models (LLMs). It focuses on assessing the models’ performance in handling extended text inputs in both English and Chinese. The benchmark includes 21 datasets covering six task categories: single-document QA, multi-document QA, summarization, few-shot learning, synthetic tasks, and code completion.

\begin{figure}
\centering
\includegraphics[width=1\linewidth]{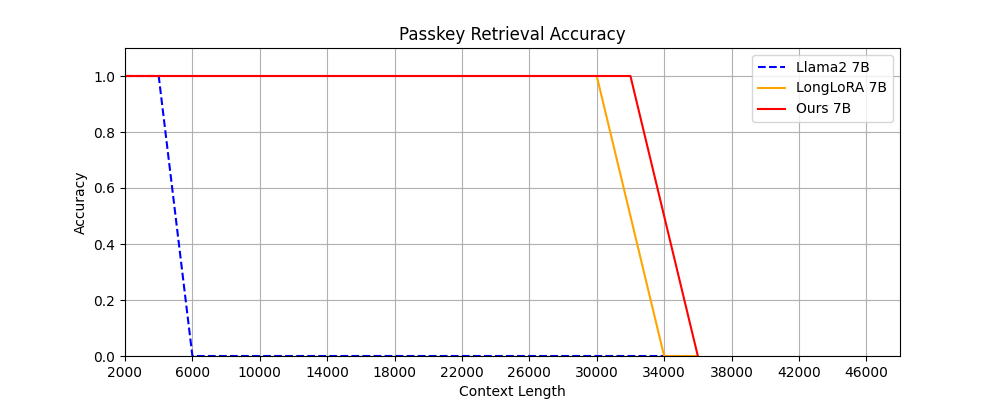}
\caption{\label{fig:figure5}Accuracy comparison on passkey retrieval between Llama2 7B and our 7B model fine-tuned on a context length of 32,768. Our model shows no retrieval accuracy degradation up to 33k or 36k, surpassing the context length, compared to LongLoRA which is 30k or 34k. }
\end{figure}

\begin{figure}
\centering
\includegraphics[width=0.6\linewidth]{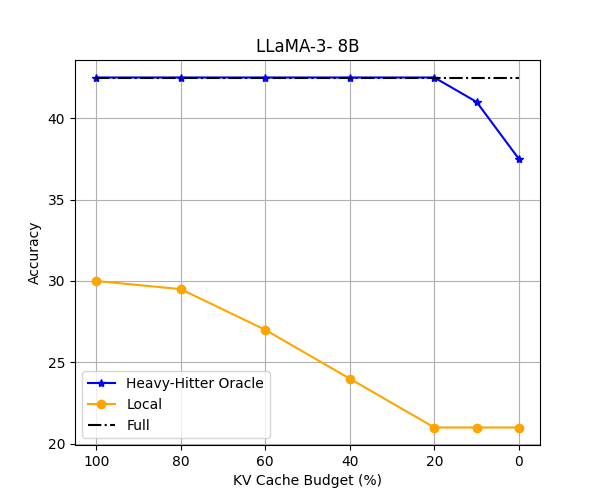}
\caption{\label{fig:figure6}Comparison of results between the baseline model with full cache, the H$_2$O method \cite{zhang2024h2o}, and the "Local" strategy that utilizes the most recent KV embeddings. The test dataset is OpenBookQA \cite{mihaylov2018can}, and the model used is Llama-3-8B \cite{meta2024llama3}.}
\end{figure}

\begin{table}[H]
    \centering
    \caption{Inference hours for LLaMA-3 8B with different KV Cache Budgets. "Full" indicates the absence of any static sparsity method. "H$_2$O" refers to the Heavy Hitter Oracle KV cache compression method. The input file for inference is the validation split of PG-19 \cite{rae2019compressive}.}
    \vspace{0.5cm}

    \begin{tabular}{lccc}
        \toprule
        \textbf{Static Sparsity} & \multicolumn{3}{c}{\textbf{KV Cache Budget (\%)}} \\
        \cmidrule(lr){2-4}
        & \textbf{50} & \textbf{75} & \textbf{100} \\
        \midrule
        Full & 9.2 & 6.9 & 4.7 \\
        H$_2$O & 6.2 & 5.8 & 4.4 \\
        \bottomrule
    \end{tabular}
    \label{tab:table5}
\end{table}

\begin{table}[H]
\centering
\caption{Performance comparison across various models and tasks. Evaluation on the LongBench \cite{bai2023longbench} benchmark. In each column, the highest value is highlighted in bold and the second highest value is underlined.}
\vspace{0.5cm}
\resizebox{\textwidth}{!}{
\begin{tabular}{lccccccc}
\toprule
Model & Avg & Single-Doc QA & Multi-Doc QA & Summarization & Few-shot Learning & Code & Synthetic \\
\midrule
GPT-3.5-Turbo & \textbf{44.0} & \textbf{39.8} & \textbf{38.7} & 26.5 & \textbf{67.1} & 54.1 & \textbf{37.8} \\
\midrule
Llama2-7B-chat & 31.0 & 24.9 & 22.6 & 24.7 & 60.0 & 48.1 & 5.9 \\
LongChat-v1.5-7B & 34.3 & 28.7 & 20.6 & 26.7 & 60.0 & 54.1 & 15.8 \\
Vicuna-v1.5-7B & 31.9 & 28.0 & 18.6 & 26.0 & 66.2 & 47.3 & 5.5 \\
LongLoRA-7B & 36.8 & 28.7 & 28.1 & 27.8 & 63.7 & 56.0 & 16.7 \\

\midrule
Ours-7B & 38.8 & 35.9 & 32.7 & \textbf{31.1} & 65.4 & \textbf{59.2} & 22.3 \\
\bottomrule
\end{tabular}}
\label{tab:table6}
\end{table}

\subsection{Main Results}

\textbf{Perplexity Result}.

Analyzing the data from Table \textcolor{red}{\ref{tab:table2}} and Table \textcolor{red}{\ref{tab:table3}}, we observe the following:

Table \textcolor{red}{\ref{tab:table2}} presents the perplexity evaluation on the proof-pile test split using three attention mechanisms: Full-Attn, S$^2$-Attn, and SF-Attn for 7B and 8B model sizes. The 7B-Llama2 model, with a training context length of 8192, shows that the SF-Attn mechanism consistently outperforms the other attention mechanisms across all target context lengths (4096, 6144, and 8192). Specifically, the perplexity values for SF-Attn are 8.60, 8.17, and 7.85, respectively, indicating more efficient performance.

Similarly, the 8B-Llama3 model, trained with the same context length, exhibits lower perplexity scores when using the SF-Attn mechanism: 5.11, 4.86, and 4.66 for target context lengths of 4096, 6144, and 8192, respectively. This demonstrates that the SF-Attn mechanism provides a notable improvement in perplexity over both Full-Attn and S$^2$-Attn.

Table \textcolor{red}{\ref{tab:table3}} summarizes the maximum context length we can fine-tune for various model sizes using a single 2x A100 machine. Both Llama2 and Llama3 models were evaluated using the same training settings. The 7B-Llama2 model shows perplexity results of 8.73, 8.55, 8.30, 8.13, and 8.05 for evaluation context lengths of 2048, 4096, 8192, 12288, and 16384, respectively. The 8B-Llama3 model achieves perplexity values of 5.11, 5.02, 4.89, 4.77, and 4.72 for the same evaluation context lengths.

These results indicate that the SF-Attn mechanism outperforms both Full-Attn and S$^2$-Attn mechanisms in terms of perplexity, especially for larger context lengths. This validates the efficiency and effectiveness of the H$_2$O algorithm within the LongLoRA framework.

\textbf{Passkey Retrieval Result}.

Analyzing the data from Table \textcolor{red}{\ref{tab:table4}} and Figure \textcolor{red}{\ref{fig:figure5}}, we observe the following:

Table \textcolor{red}{\ref{tab:table4}} presents the topic retrieval evaluation with LongChat. This task involves retrieving target topics from very long conversations with context lengths of 3k, 6k, 10k, 13k, and 16k. Our model, fine-tuned on Llama3 8B, achieves comparable performance to the state-of-the-art LongChat-13B with a lower fine-tuning cost. Specifically, our model maintains an accuracy of 1.00 for context lengths up to 10k and shows a slight decrease to 0.98 and 0.96 for 13k and 16k context lengths, respectively. This demonstrates our model's robustness and efficiency in handling long-context retrieval tasks.

Figure \textcolor{red}{\ref{fig:figure5}} compares the passkey retrieval accuracy between Llama2 7B, LongLoRA 7B, and our 7B model fine-tuned on a context length of 32,768. Our model shows no retrieval accuracy degradation up to 33k or 36k, surpassing the context length limits of LongLoRA, which are 30k and 34k. This indicates that our model can handle longer context lengths with higher accuracy compared to existing models.

\textbf{LongBench Result}.

Analyzing the data from Table \textcolor{red}{\ref{tab:table6}}, we observe  that GPT-3.5-Turbo consistently outperforms other models across various tasks, achieving the highest average score (44.0). This indicates its superior overall performance. Notably, our model, Ours-7B, secures the second highest average score (38.8), demonstrating competitive performance. Specifically, Ours-7B excels in the Code task with a leading score of 59.2, surpassing GPT-3.5-Turbo (54.1).

Despite these strengths, there are areas for improvement, particularly in the Synthetic task where Ours-7B scores 22.3, significantly lower than GPT-3.5-Turbo’s 37.8. This highlights the need to enhance our model’s capabilities in synthetic tasks to further boost its overall performance. Overall, while GPT-3.5-Turbo remains the top performer, our model shows promising results and potential for optimization.

\textbf{H$_2$O Inference Result}

In this analysis, we examine the performance and efficiency of various KV cache methods for the LLaMA-3-8B model, as depicted in Figure \ref{fig:figure5} and Table \ref{tab:table6}.

Figure \ref{fig:figure5} presents a comparative analysis between the baseline model utilizing a full cache, the H$_2$O method, and the “Local” strategy, which employs the most recent KV embeddings. The evaluation was conducted on the OpenBookQA dataset. Notably, the Heavy-Hitter Oracle (H$_2$O) method sustains high accuracy levels even with a reduced KV cache budget, demonstrating its effectiveness in preserving model performance. Conversely, the Local method exhibits a marked decline in accuracy as the KV cache budget diminishes.

Table \ref{tab:table6} details the inference hours required for LLaMA-3-8B under different KV cache budgets. The H$_2$O method substantially reduces inference time compared to the Full method across all cache budgets. For instance, at a 100\% KV cache budget, the H$_2$O method decreases inference time from 4.7 hours (Full) to 4.4 hours. This reduction is even more significant at lower KV cache budgets, underscoring the efficiency of the H$_2$O method in environments with limited resources.

In conclusion, the H$_2$O method provides an optimal balance between maintaining high accuracy and reducing inference time, thereby representing a valuable strategy for optimizing large language models such as LLaMA-3-8B. This method’s ability to offer substantial computational savings while preserving performance highlights its potential for broader applications in resource-constrained settings.

\begin{table}[H]
    \centering
    \caption{Perplexity of Different Training Methods. The results are based on the same experimental setup as Table \textcolor{red}{\ref{tab:table2}}, using the validation split of the PG-19 dataset \cite{rae2019compressive}. The comparison includes SF-Attn, S$^2$-Attn with Global Attention, S$^2$-Attn with the Segmentation and Reassembly (S\&R) Algorithm, and S$^2$-Attn alone.}
    \vspace{0.5cm}
    \begin{tabular}{lc}
        \toprule
        \textbf{Training Method} & \textbf{Perplexity} \\
        \midrule
        SF-Attn & 8.64 \\
        S$^2$-Attn + Global Attention & \textbf{8.91} \\
        S$^2$-Attn + S\&R Algorithm & \textbf{8.86} \\
        S$^2$-Attn & 9.09 \\
        \bottomrule
    \end{tabular}
    \label{tab:table7}
\end{table}

\begin{table}[H]
    \centering
    \caption{Perplexity results of ablation experiments on different training methods. The experiments were conducted using the same setup as Table \textcolor{red}{\ref{tab:table2}}, utilizing the validation split of the PG-19 dataset \cite{rae2019compressive}. The comparison includes S$^2$Attn, shifting up only, shifting back only, and S$^2$-Attn with the Segmentation and Reassembly (S\&R) Algorithm. }
    \vspace{0.5cm}
    \begin{tabular}{lc}
        \toprule
        \textbf{Training Method} & \textbf{Perplexity} \\
        \midrule
        S$^2$Attn & 9.09 \\
        Just shift up & \textbf{9.35} \\
        Just shift back & \textbf{9.60} \\
        S$^2$-Attn + S\&R Algorithm & 8.86 \\
        \bottomrule
    \end{tabular}
    \label{tab:table8}
\end{table}

\begin{table}[H]
	\centering
	\fontsize{8}{11}\selectfont    %{字体尺寸}{行距}
	\caption{Ablation on the variants of S\(^2\)-Attn. These variants are illustrated in Figure 6. Similar to the setting in Table 7, we fine-tune a Llama2 7B to 8192 context and evaluate on PG19 validation set.}
 \vspace{0.5cm}
	\begin{tabular}{cccccc}
		\toprule
		\toprule
		\multirow{2}{*}{Setting}&\multirow{2}{*}{Traing}&\multirow{2}{*}{Attention}&
		\multicolumn{3}{c}{Target Context Length} \cr
		\cmidrule(lr){4-6}
		& & & 4096 & 6144 & 8192  \cr
		\cmidrule(lr){1-6}
		\multirow{5}{*}{7B-Llama2}
		
            && Full-Attn & 8.59 & 8.16 & 7.83   \\
		&& $S^2$-Attn & 9.09 & 8.82 & 8.64 \\
             &8192&Sparse Fixed Attention & 8.85 & 8.50 & 8.24 \\
             &&Stride Attention & 8.81 & 8.47 & 8.22 \\
             &&Random Attention & 8.78 & 8.44 & 8.20 \\
		&&SF-Attn &\textbf{8.60} & \textbf{8.17 }& \textbf{7.85}  \\

		\bottomrule
		\bottomrule
	\end{tabular}\vspace{0cm}
	\label{tab:table9}
\end{table}

\subsection{Ablation Study}

\textbf{Ablation on SF-Attn training steps}

To evaluate the effectiveness of different components in our SF-Attn training methodology, we conducted a series of ablation studies. Table \textcolor{red}{\ref{tab:table7}}  presents the perplexity results for various training methods using the validation split of the PG-19 dataset. The SF-Attn method alone yields a perplexity of 8.64, establishing a baseline for comparison. Introducing global attention to S\textsuperscript{2}-Attn results in a perplexity of 8.91, indicating that global attention alone does not significantly improve performance. Combining S\textsuperscript{2}-Attn with the Segmentation and Reassembly (S\&R) Algorithm results in a perplexity of 8.86, showing a moderate improvement over S\textsuperscript{2}-Attn alone. The baseline S\textsuperscript{2}-Attn method alone has a higher perplexity of 9.09 compared to the SF-Attn baseline, demonstrating the need for additional enhancements to achieve better performance. These results indicate that while the SF-Attn method alone is effective, combining it with additional techniques like the S\&R Algorithm can yield improvements in perplexity.

Further ablation studies were conducted to analyze the impact of individual components of the Segmentation and Reassembly (S\&R) Algorithm. Table \textcolor{red}{\ref{tab:table8}} provides perplexity results for these experiments. The baseline S\textsuperscript{2}-Attn method has a perplexity of 9.09. When the model only applies the "shift up" step, the perplexity increases to 9.35, indicating that this step alone is not sufficient for improving performance. Applying only the "shift back" step results in a perplexity of 9.60345459, showing that this step alone is also insufficient. The complete Segmentation and Reassembly Algorithm achieves a perplexity of 8.86, demonstrating the effectiveness of combining both shifting steps. These ablation studies highlight the importance of the Segmentation and Reassembly Algorithm in enhancing the performance of the SF-Attn training method. By integrating both the "shift up" and "shift back" steps, the S\&R Algorithm effectively reduces perplexity and improves model performance.

\textbf{Ablation on SF-Attn Varation}

specifically focusing on the Sparse Fixed Attention (SF-Attn) and its variations. Figure \ref{fig:figure7} illustrates the different attention patterns: Sparse Fixed Attention, Stride Attention, and Random Attention. These variations represent different strategies for distributing attention across the token sequence. Sparse Fixed Attention uses a fixed sparse pattern, Stride Attention distributes attention in a stride pattern, and Random Attention allocates attention randomly.

Table \ref{tab:table9} presents the results of fine-tuning a Llama2 7B model on sequences with varying target context lengths (4096, 6144, and 8192) and evaluating on the PG19 validation set. The performance is measured across different settings of the attention mechanism. Key observations include that Full Attention (Full-Attn) shows consistent performance but is computationally expensive. S²-Attention (S²-Attn) performs slightly better than Full Attention, particularly at longer context lengths. Sparse Fixed Attention maintains competitive performance and is more efficient. Stride and Random Attention also perform well but slightly lower than Sparse Fixed Attention. The SF-Attention method, combining elements of these strategies, achieves the best balance with the lowest perplexity scores, indicating it is the most efficient and effective attention mechanism for handling long sequences.

\section{Discussion}

\subsection{The Failure of Shift Operation}

The shift operation of Sparse Shift Attention \cite{chen2023longlora} is inspired by the Swin Transformer \cite{liu2021swin}. It aims to facilitate more information exchange between different groups of tokens. However, this approach also leads to an information leaking problem.

The information leaking issue likely arises because the shift operation allows tokens from different segments to share information too freely. While the intention is to enhance the model’s ability to integrate information from various parts of the sequence, it inadvertently causes tokens to access information that they should not be aware of at that particular stage of processing. This premature exposure to information can disrupt the model’s learning process, leading to overfitting or incorrect associations.

Moreover, the shift operation might blur the distinct boundaries between token groups, causing the model to lose track of the specific context within each group. This can result in a degradation of performance, as the model might struggle to maintain a coherent understanding of the local context, which is essential for accurately interpreting and generating long sequences.

In summary, while the shift operation aims to improve information exchange, it inadvertently causes information leakage by allowing tokens to prematurely access and integrate information from different segments. This undermines the model’s ability to maintain distinct contextual boundaries and can negatively impact its performance on tasks requiring precise contextual understanding.

\subsection{The Success of Making Sink Attention Tokens Globally Operative}

The method described in the following section details the successful implementation of globally operative sink attention tokens, enhancing the model’s ability to handle long sequences efficiently.

The successful results observed from implementing sink attention tokens can be attributed to several key factors. First, by designating specific tokens as sink attention tokens that attend to all other tokens in the sequence, the model can effectively capture and integrate information from across the entire sequence. This global attention mechanism ensures that critical information is not lost, even in very long sequences.

Furthermore, by ensuring that all tokens in the sequence also attend to the sink attention tokens, the model can maintain a coherent understanding of the sequence context. This bidirectional attention flow allows the model to reinforce important information at multiple stages, enhancing the overall comprehension and retention of context.

The efficient complexity of  O(n log n)  achieved through this method, due to the relatively small number of sink attention tokens, ensures that this enhanced capability does not come at the cost of significantly increased computational overhead. This balance between maintaining comprehensive attention and computational efficiency is likely a key factor in the improved performance observed in models utilizing this technique.

In summary, the implementation of globally operative sink attention tokens allows the model to maintain a robust and coherent understanding of long sequences, ensuring critical information is attended to throughout the processing stages, thus leading to improved performance on tasks requiring extensive context comprehension.

\subsection{The Convenience of Applying KV Cache Compression Function}

As the results shown in the H$_2$O inference test indicate, the H$_2$O method can maintain accuracy even when the KV cache budget is reduced by half. However, the time savings achieved with this method are significant, reducing inference time by a factor of 1.5. This capability allows for flexible deployment of inference methods based on the available computing resources. The ability to adjust the KV cache budget without compromising accuracy ensures that large language models can be efficiently and effectively utilized in resource-constrained environments, optimizing both performance and computational efficiency.

\subsection{The Good Result of Chatbot}
As shown in the example of chat ability comparison in Section \ref{sec:C}, we observe the responses from two versions of the LongLoRA model to a story-related question.

The user input describes a narrative involving an old fisherman named Tom, who ventures farther out to sea to a place known as the “Blue Deep” in hopes of finding fish after several unsuccessful weeks. The question asks why Tom decided to venture farther out to sea than he ever had before.

The LongLoRA-7B model’s response focuses on Tom’s desperation and worry about not catching any fish, emphasizing his repeated return with an empty net and his determination to support his family. This response accurately captures the key elements of Tom’s motivation as described in the narrative.

On the other hand, the SinkLoRA-7B model provides a \textbf{more detailed} explanation, highlighting the prolonged disappearance of fish and Tom’s resultant decision to take the risk of venturing to the “Blue Deep.” It also mentions Tom’s hope to find fish there, providing a more comprehensive understanding of his motivations.

The reason for the differing levels of detail and context in the responses can be attributed to the training and fine-tuning differences between the two models. The SinkLoRA-7B model might have been trained on a more diverse dataset or undergone additional fine-tuning to better understand and generate contextually rich responses. This additional training could enable it to provide more nuanced and detailed answers, capturing the subtleties of the narrative more effectively.

In conclusion, both models successfully identify Tom’s primary motivation, but the SinkLoRA-7B model offers a more thorough and contextually rich explanation. This comparison underscores the effectiveness of the chat capabilities in understanding and accurately responding to narrative-based questions. The observed differences highlight the importance of extensive training and fine-tuning in enhancing model performance and response quality.
% \subsection{the good result of impying a larger instructed dataset}

\section{Conclusion and Future Work}
In this study, we introduced SinkLoRA, a significant enhancement over the original LongLoRA, designed to improve the efficiency and performance of large language models (LLMs) in handling long-context sequences. SinkLoRA addresses the limitations of the previous model by implementing Sink Fixed Attention (SF-Attn) and utilizing advanced KV cache compression techniques like the Heavy-Hitter Oracle (H2O). Our proposed SF-Attn method effectively redistributes attention scores, reducing the overemphasis on initial tokens and improving overall model accuracy. This approach, combined with the segmentation and reassembly algorithm, allows for better handling of extended contexts without increasing computational complexity. The integration of the H2O KV cache compression further accelerates inference, making SinkLoRA a highly efficient solution for deploying LLMs in resource-constrained environments.

Future work will focus on further optimizing the attention mechanisms and exploring the compatibility of SinkLoRA with other types of LLMs and position encodings. We also plan to investigate more advanced KV cache management techniques to enhance the flexibility and efficiency of inference processes. The goal is to continue improving the performance and scalability of LLMs, enabling their application in a broader range of tasks and environments. In summary, SinkLoRA represents a substantial step forward in the development of efficient long-context processing techniques for large language models, offering promising avenues for future research and application.

\bibliographystyle{plain}
\bibliography{neurips_2024.bib}

\newpage

% \textcolor{red}{}

\appendix

\section*{Appendix}
We organize our supplementary material as follows.

\begin{itemize}
    \item  Section \ref{sec:A} 
    \item  Section \ref{sec:B}
    \item  Section \ref{sec:C}
%     \item In Section \ref{sec:D}, we show the effectiveness of our training strategy by comparing with sandwich sampling rule and inplace distillation \cite{ref63}.
%     \item In Section \ref{sec:E}, we discuss the effect of training without the pretrained weights of anchors.
%     \item In Section \ref{sec:F}, we experiment with different number of samples for initializing stitching layers.
%     \item In Section \ref{sec:G}, we provide additional discussion with One-shot NAS.
%     \item In Section \ref{sec:H}, we compare SN-Net with Layer-Drop \cite{ref12} at inference time.
\end{itemize}

\section{Varation of SF-Attn}\label{sec:A}

\begin{figure}
\centering
\includegraphics[width=1\linewidth]{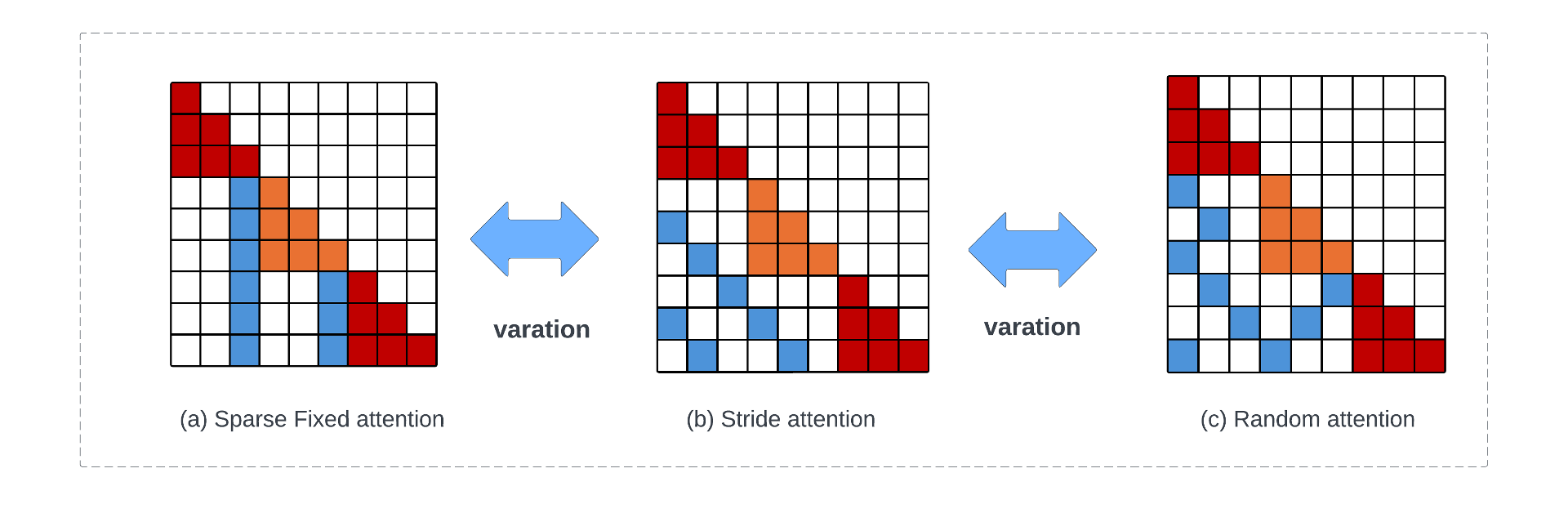}
\caption{\label{fig:figure7}Illustration of different attention mechanisms utilized in the model: (a) Sparse Fixed attention, where attention is fixed to certain positions creating a sparse matrix, (b) Stride attention, where attention is distributed in a stride pattern allowing broader context capture, and (c) Random attention, where attention is allocated randomly across the sequence, enabling the model to focus on diverse parts of the input. These variations demonstrate different strategies for distributing attention in the model’s architecture to enhance performance and context understanding.}
\end{figure}

We provide Varation of SF-Attn in Figure 7.

\section{Dataset Overview}\label{sec:B}

% the \\ insures the section title is centered below the phrase: Appendix B

We provide Overview of Dataset in Figure 8.
\begin{figure}
\centering
\includegraphics[width=1.0\linewidth]{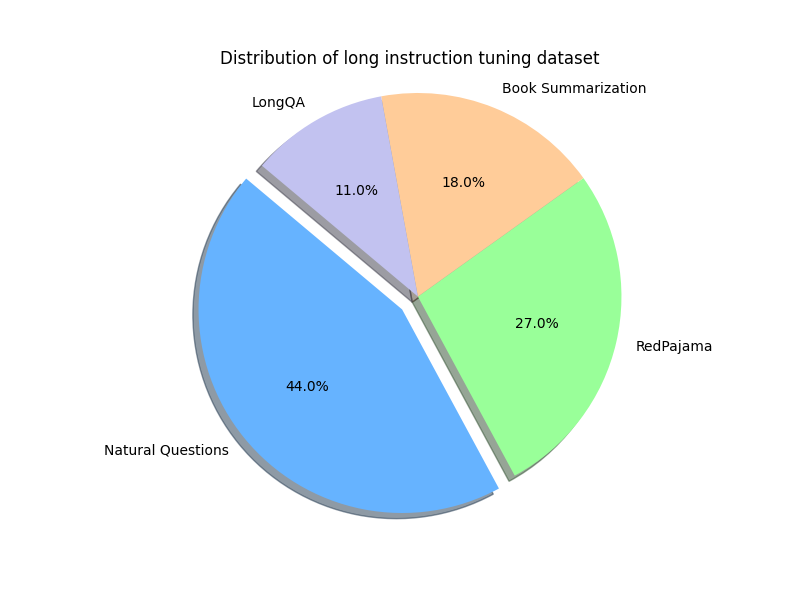}
\caption{\label{fig:figure8}Overview of LongAlpaca-plus}
\end{figure}

\section{Example of Chat}\label{sec:C}

We provide a example of dialogue in Figure 9.

\begin{figure}
\centering
\includegraphics[width=1\linewidth]{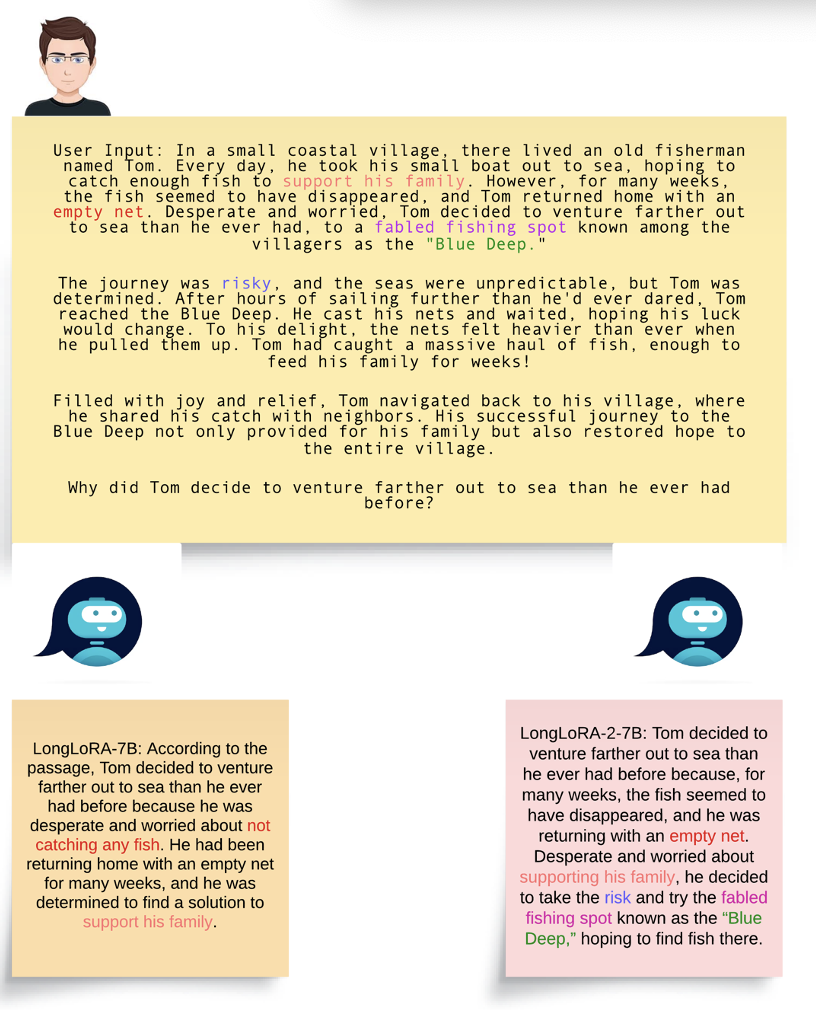}
\caption{\label{fig:figure9}Examples on story-related questions. We compare the answers from our 7B model with SFT and the chat version of LongLoRA-7B.
}
\end{figure}

\end{document}